\documentclass[english]{article}
\usepackage[latin9]{inputenc}
\usepackage[letterpaper]{geometry}
\usepackage{babel}
\usepackage{amsmath}
\usepackage{amssymb}
\usepackage{graphicx}
\usepackage{esint}
\usepackage[unicode=true,pdfusetitle,
 bookmarks=true,bookmarksnumbered=false,bookmarksopen=false,
 breaklinks=false,pdfborder={0 0 1},backref=section,colorlinks=false]
 {hyperref}

\makeatletter

\providecommand{\tabularnewline}{\\}

\usepackage{babel}
\usepackage{amsbsy}

\usepackage{flushend}

\usepackage{subfigure}

\usepackage{algorithm}\usepackage{algorithmic}

\usepackage{babel}
\date{}

\makeatother

\begin{document}

\title{Thurstonian Boltzmann Machines: Learning from Multiple Inequalities}

\author{Truyen Tran$^{\dagger\ddagger}$, Dinh Phung$^{\dagger}$, Svetha
Venkatesh$^{\dagger}$ \\
 $^{\dagger}$Pattern Recognition and Data Analytics, Deakin University,
Australia\\
 $^{\ddagger}$Department of Computing, Curtin University, Australia\\
 \{truyen.tran,dinh.phung,svetha.venkatesh\}@deakin.edu.au }

\maketitle
\global\long\def\BigO{\mathcal{O}}
 \global\long\def\U{\mathcal{U}}
 \global\long\def\I{\mathcal{I}}
 \global\long\def\M{\mathcal{M}}
 \global\long\def\B{\mathcal{B}}
 \global\long\def\Loss{\mathcal{L}}
 \global\long\def\userset{\mathcal{U}}
 \global\long\def\itemset{\mathcal{I}}
 \global\long\def\ratemat{\mathcal{R}}
 \global\long\def\A{\mathcal{A}}

\global\long\def\rateset{\mathcal{S}}
 \global\long\def\hb{\boldsymbol{h}}
 \global\long\def\h{h}
 \global\long\def\xb{\boldsymbol{x}}
 \global\long\def\x{x}
 \global\long\def\eb{\boldsymbol{e}}
 \global\long\def\e{e}

\global\long\def\wb{\boldsymbol{w}}
 \global\long\def\w{w}
 \global\long\def\qb{\boldsymbol{q}}
 \global\long\def\q{q}
 \global\long\def\pb{\boldsymbol{p}}
 \global\long\def\p{p}
 \global\long\def\gb{\boldsymbol{g}}
 \global\long\def\ab{\boldsymbol{a}}
 \global\long\def\bb{\boldsymbol{b}}
 \global\long\def\cb{\boldsymbol{c}}
 \global\long\def\db{\boldsymbol{d}}
 \global\long\def\vb{\boldsymbol{v}}

\global\long\def\thetab{\boldsymbol{\theta}}
 \global\long\def\alphab{\boldsymbol{\alpha}}
 \global\long\def\etab{\boldsymbol{\eta}}
 \global\long\def\nub{\boldsymbol{\nu}}
 \global\long\def\betab{\boldsymbol{\beta}}
 \global\long\def\mub{\boldsymbol{\mu}}
 \global\long\def\gammab{\boldsymbol{\gamma}}

\global\long\def\deltab{\boldsymbol{\delta}}
 \global\long\def\xib{\boldsymbol{\xi}}
 \global\long\def\taub{\boldsymbol{\tau}}
 \global\long\def\lambdab{\boldsymbol{\lambda}}
 \global\long\def\LL{\mathcal{L}}
 \global\long\def\Domain{\Omega}
 \global\long\def\Domainb{\boldsymbol{\Omega}}

\global\long\def\Real{\mathbb{R}}
 \global\long\def\Id{\mathbb{I}}
 \global\long\def\Normal{\mathcal{N}}
 \global\long\def\Model{\mathrm{TBM}}
 \global\long\def\Ib{\boldsymbol{I}}

\global\long\def\hb{\boldsymbol{h}}
 \global\long\def\h{h}
 \global\long\def\xb{\boldsymbol{x}}
 \global\long\def\x{x}
 \global\long\def\eb{\boldsymbol{e}}
 \global\long\def\e{e}
 \global\long\def\vb{\boldsymbol{e}}
 \global\long\def\ub{\boldsymbol{x}}
 \global\long\def\v{e}
 \global\long\def\u{x}

\global\long\def\LL{\mathcal{L}}
 \global\long\def\Domain{\Omega}
 \global\long\def\Domainb{\boldsymbol{\Omega}}

\global\long\def\wb{\boldsymbol{w}}
 \global\long\def\w{w}
 \global\long\def\qb{\boldsymbol{q}}
 \global\long\def\q{q}
 \global\long\def\pb{\boldsymbol{p}}
 \global\long\def\p{p}
 \global\long\def\gb{\boldsymbol{g}}
 \global\long\def\ab{\boldsymbol{a}}
 \global\long\def\bb{\boldsymbol{b}}
 \global\long\def\cb{\boldsymbol{c}}
 \global\long\def\db{\boldsymbol{d}}

\global\long\def\thetab{\boldsymbol{\theta}}
 \global\long\def\alphab{\boldsymbol{\alpha}}
 \global\long\def\etab{\boldsymbol{\eta}}
 \global\long\def\nub{\boldsymbol{\nu}}
 \global\long\def\betab{\boldsymbol{\beta}}
 \global\long\def\mub{\boldsymbol{\mu}}
 \global\long\def\gammab{\boldsymbol{\gamma}}

\begin{abstract}
We introduce \emph{Thurstonian Boltzmann Machines} ($\Model$), a
unified architecture that can naturally incorporate a wide range of
data inputs at the same time. Our motivation rests in the Thurstonian
view that many discrete data types can be considered as being generated
from a subset of underlying latent continuous variables, and in the
observation that each realisation of a discrete type imposes certain
inequalities on those variables. Thus learning and inference in $\Model$
reduce to making sense of a set of inequalities. Our proposed $\Model$
naturally supports the following types: Gaussian, intervals, censored,
binary, categorical, muticategorical, ordinal, (in)-complete rank
with and without ties. We demonstrate the versatility and capacity
of the proposed model on three applications of very different natures;
namely handwritten digit recognition, collaborative filtering and
complex social survey analysis. 
\end{abstract}

\section{Introduction}

Restricted Boltzmann machines (RBMs) have proved to be a versatile
tool for a wide variety of machine learning tasks and as a building
block for deep architectures \cite{hinton2006rdd,salakhutdinov2009deep,smolensky1986information}.
The original proposals mainly handle binary visible and hidden units.
Whilst binary hidden units are broadly applicable as feature detectors,
non-binary visible data requires different designs. Recent extensions
to other data types result in type-dependent models: the Gaussian
for continuous inputs \cite{hinton2006rdd}, Beta for bounded continuous
inputs \cite{le2011learning}, Poisson for count data \cite{gehler2006rap},
multinomial for unordered categories \cite{salakhutdinov2009replicated},
and ordinal models for ordered categories \cite{Truyen:2009a,Truyen:2012d}.

The Boltzmann distribution permits several types to be jointly modelled,
thus making the RBM a good tool for multimodal and complex social
survey analysis. The work of \cite{ngiam2011multimodal,srivastava2012multimodal,xing2005mining}
combines continuous (e.g., visual and audio) and discrete modalities
(e.g., words). The work of \cite{Truyen:2011b} extends the idea further
to incorporate ordinal and rank data. However, there are conceptual
drawbacks: First, conditioned on the hidden layer, they are still
separate type-specific models; second, handling ordered categories
and ranks is not natural; and third, specifying direct correlation
between these types remains difficult.

The main thesis of this paper is that many data types can be captured
in one unified model. The key observations are that (i) type-specific
properties can be modelled using one or several \emph{underlying continuous
variables}, in the spirit of Thurstonian models%
\footnote{Whilst Thurstonian models often refer to human's judgment of discrete
choices, we use the term ``Thurstonian'' more freely without the
notion of human's decision.%
} \cite{thurstone1927law}, and (ii) evidences be expressed in the
form of one or several \emph{inequalities} of these underlying variables.
For example, a binary visible unit is turned on if the underlying
variable is beyond a threshold; and a category is chosen if its utility
is the largest among all those of competing categories. The use of
underlying variables is desirable when we want to explicitly model
the generative mechanism of the data. In psychology and economics,
for example, it gives much better interpretation on why a particular
choice is made given the perceived utilities \cite{bockenholt2006thurstonian}.
Further, it is natural to model the correlation among type-specific
inputs using a covariance structure on the underlying variables.

The inequality observation is interesting in its own right: Instead
of learning from assigned values, we learn from the \emph{inequality
expression of evidences}, which can be much more relaxed than the
value assignments. This class of evidences indeed covers a wide range
of practical situations, many of which have not been studied in the
context of Boltzmann machines, as we shall see throughout the paper.

To this end, we propose a novel class of models called \emph{Thurstonian
Boltzmann Machine} ($\Model$). The $\Model$ utilises the Gaussian
restricted Boltzmann machine (GRBM): The top layer consists of binary
hidden units as in standard RBMs; the bottom layer contains a collection
of Gaussian variable groups, one per input type. The main difference
is that $\Model$ does not require valued assignments for the bottom
layer \emph{but a set of inequalities expressing the constraints imposed
by the evidences}. Except for a limiting case of point assignments
where the inequalities are strictly equalities, the Gaussian layer
is never fully observed. The $\Model$ supports more data types in
a unified manner than ever before: For any combination of the point
assignments, intervals, censored values, binary, unordered categories,
multi-categories, ordered categories, (in)-complete ranks with and
without ties, all we need to do is to supply relevant subset of inequalities.

We evaluate the proposed model on three applications of very different
natures: handwritten digit recognitions, collaborative filtering and
complex survey analysis. For the first two applications, the performance
is competitive against methods designed for those data types. On the
last application, we believe we are among the first to propose a scalable
and generic machinery for handle those complex data types.

\section{Gaussian RBM}

Let $\xb=(x_{1},x_{2},...,x_{N})^{\top}\in\Real^{N}$ be a vector
of input variables. Let $\hb=(h_{1},h_{2},...,h_{K})^{\top}\in\left\{ 0,1\right\} ^{K}$
be a set of hidden factors which are designed to capture the variations
in the observations. The input layer and the hidden layer form an
undirected bipartite graph, i.e., only cross-layer connections are
allowed. The model admits the Boltzmann distribution\vspace{-0.1cm}

\begin{align}
P(\xb,\hb) & =\frac{1}{Z}\exp\left\{ -E(\xb,\hb)\right\} \label{eq:Model-def}
\end{align}
\vspace{-0.1cm}
 where $Z=\sum_{\hb}\int\exp\left\{ -E(\xb,\hb)\right\} d\xb$ is
the normalising constant and $E(\xb,\hb)$ is the state energy. The
energy is decomposed as\vspace{-0.1cm}

\begin{equation}
E(\xb,\hb)=\sum_{i}\left(\frac{\x_{i}^{2}}{2}-\left(\alpha_{i}+W_{i\bullet}\hb\right)x_{i}\right)-\gammab^{\top}\hb\label{eq:energy-def}
\end{equation}
\vspace{-0.1cm}
 where$\left\{ \alpha_{i}\right\} _{i=1}^{N},W=\{W_{ik}\},\gammab=\{\gamma_{k}\}$
are free parameters and $W_{i\bullet}$ denotes the $i$-th row.

Given the input $\xb$, the posterior has a simple form\vspace{-0.1cm}

\begin{eqnarray}
P(\hb\mid\xb) & = & \prod_{k}P(h_{k}\mid\xb)\label{eq:posterior-factor}\\
P(h_{k}=1\mid\xb) & = & \frac{1}{1+e^{-\gamma_{k}-W_{\bullet k}^{'}\xb}}\nonumber 
\end{eqnarray}
\vspace{-0.1cm}
 where $W_{\bullet k}$ denotes the $k$-th column. Similarly, the
generative process given the binary factor $\hb$ is also factorisable\vspace{-0.1cm}

\begin{eqnarray}
P(\xb\mid\hb) & = & \prod_{i}P(\x_{i}\mid\hb)\label{eq:generative-factor}\\
P(\x_{i}\mid\hb) & = & \Normal(\alpha_{i}+W_{i\bullet}\hb,1)\nonumber 
\end{eqnarray}
\vspace{-0.1cm}
 where $\Normal(\mu,1)$ is the normal distribution of mean $\mu$
and unit deviation.

\section{Thurstonian Boltzmann Machines\label{sec:Probit BM}}

We now generalise the Gaussian RBM into the Thurstonian Boltzmann
Machine ($\Model$). Denote by $\eb$ an observed \emph{evidence}
of $\xb$. Standard evidences are the point assignment of $\xb$ to
some specific real-valued vector, i.e., $\xb=\vb$. Generalised evidences
can be expressed using \emph{inequality constraints} 
\begin{equation}
\bb\boldsymbol{\leq}_{\odot}A\xb\boldsymbol{\le}_{\odot}\cb\label{eq:ineq-constraint}
\end{equation}
for some transform matrix $A\in\Real^{M\times N}$ and vectors $\bb,\cb\in\Real^{M}$,
where $\boldsymbol{\le}_{\odot}$ denotes element-wise inequalities.
Thus an evidence can be completely realised by specifying the triple
$\left\langle A,\bb,\cb\right\rangle $. For example, for the point
assignment, $\left\langle A=\boldsymbol{I},\bb=\vb,\cb=\vb\right\rangle $,
where $\boldsymbol{I}$ is the identity matrix. In what follows, we
will detail other useful popular realisations of these quantities.

\subsection{Boxed Constraints \label{sub:Boxed-Constraints}}

This refers to the case where input variables are independently constrained,
i.e., $A=\Ib$, and thus we need only to specify the pair $\left\langle \bb,\cb\right\rangle $.

\paragraph*{Censored observations.}

This refers to situation where we only know the continuous observation
beyond a certain point, i.e., $\bb=\eb$ and $\cb=+\infty$. For example,
in survival analysis, the life expectancy of a person might be observed
up to a certain age, and we have no further information afterward.

\paragraph*{Interval observations\emph{.} }

When the measurements are imprecise, it may be better to specify the
range of possible observations with greater confidence rather than
a singe point, i.e., $\bb=\eb-\deltab$ and $\cb=\eb+\deltab$ for
some pair $(\eb,\deltab)$. For instance, missile tracking may estimate
the position of the target with certain precision.

\paragraph*{Binary observations\emph{.}}

A binary observation $e_{i}$ can be thought as a result of clipping
$x_{i}$ by a threshold $\theta_{i}$, that is $e_{i}=1$ if $x_{i}\ge\theta_{i}$
and $e_{i}=0$ otherwise. The boundaries in Eq.~(\ref{sub:Inequality-Constraints})
become: 
\begin{equation}
\left\langle b_{i},c_{i}\right\rangle =\begin{cases}
\left\langle -\infty,\theta_{i}\right\rangle  & \quad e_{i}=0\\
\left\langle \theta_{i},+\infty\right\rangle  & \quad e_{i}=1
\end{cases}\label{eq:binary-model}
\end{equation}
Thus, this model offers an alternative%
\footnote{To be consistent with the statistical literature, we can call it the
\emph{probit RBM}, which we will study in Section~\ref{sub:Probit-RBM-for-Handwritten-Digits}.%
} to standard binary RBMs of \cite{smolensky1986information,freund1994unsupervised}.

\paragraph*{Ordinal observations.}

Denote by $\eb=(\e_{1},\e_{2},...,\e_{N})$ the set of ordinal observations,
where each $e_{i}$ is drawn from an ordered set $\left\{ 1,2,..,L\right\} $.
The common assumption is that the ordinal level $e_{i}=l$ is observed
given $x_{i}\in\left[\theta_{l-1},\theta_{l}\right]$ for some thresholds
$\theta_{1}\le\theta_{2}\le...\theta_{L-1}$. The boundaries thus
read 
\begin{equation}
\left\langle b_{i},c_{i}\right\rangle =\begin{cases}
\left\langle -\infty,\theta_{1}\right\rangle  & \quad l=1\\
\left\langle \theta_{l-1},\theta_{l}\right\rangle  & \quad l=2,3,..L-1\\
\left\langle \theta_{L-1},+\infty\right\rangle  & \quad\mbox{otherwise}
\end{cases}\label{eq:ordinal-model}
\end{equation}
This offers an alternative%
\footnote{This can be called \emph{ordered probit RBM}.%
} to the ordinal RBMs of \cite{Truyen:2009a}.

\subsection{Inequality Constraints \label{sub:Inequality-Constraints}}

\paragraph*{Categorical observations\emph{.}}

This refers to the situation where out of an \emph{unordered} set
of categories, we observe only one category at a time. This can be
formulated as follows. Each category is associated with a ``utility''
variable. The category $l$ is observed (i.e., $e_{i}=m$)\emph{ }if
it has the largest utility, that is $x_{il}\ge\max_{m\ne l}x_{im}$.
Thus, $x_{il}$ is the upper-threshold for all other utilities. On
the other hand, $\max_{m\ne l}x_{im}$ is the lower-threshold for
$x_{il}$. This suggests an \emph{EM-style} procedure: (\emph{i})
fix $x_{il}$ (or treat it as a threshold) and learn the model under
the intervals $x_{im}\le x_{il}$ for all $m\ne l$, and (\emph{ii})
fix all categories other than $l$, learn the model under the interval
$x_{il}\ge\max_{m\ne l}x_{im}$. This offers an alternative%
\footnote{We can call this model \emph{multinomial probit RBM.}%
} to the multinomial logit treatment in \cite{Salakhutdinov-et-alICML07}.

To illustrate the point, suppose there are only four variables $z_{1},z_{2},z_{3},z_{4}$,
and $z_{1}$ is observed, then we have $z_{1}\ge\max\left\{ z_{2},z_{3},z_{4}\right\} $.
This can be expressed as $z_{1}-z_{2}\ge0;\quad z_{1}-z_{3}\ge0$
and $z_{1}-z_{4}\ge0$. These are equivalent to 
\[
\left\langle A=\left[\begin{array}{cccc}
1 & -1 & 0 & 0\\
1 & 0 & -1 & 0\\
1 & 0 & 0 & -1
\end{array}\right];\quad\bb=\boldsymbol{0};\quad\cb=+\infty\right\rangle 
\]

\paragraph*{Imprecise categorical observations\emph{.}}

\emph{This generalises the categorical case}: The observation is a
subset of a set, where \emph{any} member of the subset can be a possible
observation%
\footnote{This is different from saying that all the members of the subset must
be observed.%
}. For example, when asked to choose the best sport team of interest,
a person may pick two teams without saying which is the best. For
instance, suppose the subset is $\{z_{1},z_{2}\}$, then $\min\left\{ z_{1},z_{2}\right\} \ge\max\{z_{3},z_{4}\}$,
which can be expressed as $z_{1}-z_{3}\ge0;\quad z_{2}-z_{3}\ge0$,
$z_{1}-z_{4}\ge0$ and $z_{2}-z_{4}\ge0$. This translates to the
following triple 
\[
\left\langle A=\left[\begin{array}{cccc}
1 & 0 & -1 & 0\\
1 & 0 & 0 & -1\\
0 & 1 & -1 & 0\\
0 & 1 & 0 & -1
\end{array}\right];\quad\bb=\boldsymbol{0};\quad\cb=+\infty\right\rangle 
\]

\paragraph*{Rank (with Ties) observations\emph{.}}

\emph{This generalises the imprecise categorical cases}: Here we have
a (partially) ranked set of categories. Assume that the rank is produced
in a stagewise manner as follows: The best category subset is selected
out of all categories, the second best is selected out of all categories
except for the best one, and so on. Thus, at each stage we have an
imprecise categorical setting, but now the utilities of middle categories
are constrained from both sides -- the previous utilities as the upper-bound,
and the next utilities as the lower-bound.

As an illustration, suppose there are four variables $z_{1},z_{2},z_{3},z_{4}$
and a particular rank (with ties) imposes that $\min\left\{ z_{1},z_{2}\right\} \ge z_{3}\ge z_{4}$.
This be rewritten as $z_{1}\ge z_{3};\quad z_{2}\ge z_{3};\quad z_{3}\ge z_{4}$,
which is equivalent to 
\[
\left\langle A=\left[\begin{array}{cccc}
1 & 0 & -1 & 0\\
0 & 1 & -1 & 0\\
0 & 0 & 1 & -1
\end{array}\right];\quad\bb=\boldsymbol{0};\quad\cb=+\infty\right\rangle 
\]

\vspace{-0.2cm}

\section{Inference Under Linear Constraints \label{sub:Generalised-Inference}}

Under the $\Model$, MCMC-based inference without evidences is simple:
we alternate between $P(\hb\mid\xb)$ and $P(\xb\mid\hb)$. This is
efficient because of the factorisations in Eqs.~(\ref{eq:posterior-factor},\ref{eq:generative-factor}).
Inference with inequality-based evidence $\eb$ is, however, much
more involved except for the limiting case of point assignments.

Denote by $\Domainb(\eb)=\left\{ \xb\mid\bb\boldsymbol{\le}_{\odot}A\xb\boldsymbol{\le}_{\odot}\cb\right\} $
the constrained domain of $\xb$ defined by the evidence $\eb$. Now
we need to specify and sample from the constrained distribution $P(\xb,\hb\mid\eb)$
defined on $\Domainb(\eb)$. Sampling $P(\hb\mid\xb)$ remains unchanged,
and in what follows we focus on sampling from $P(\xb\mid\hb,\eb)$.

\subsection{Inference under Boxed Constraints}

For \emph{boxed constraints} (Section~\ref{sub:Boxed-Constraints}),
due to the conditional independence, we still enjoy the factorisation
$P(\xb\mid\hb,\eb)=\prod_{i}P(x_{i}\mid\hb,\eb)$. We further have
\begin{eqnarray*}
P(x_{i}\mid\hb,\eb) & = & \frac{P(x_{i}\mid\hb)}{\Phi(c_{i}\mid\hb)-\Phi(b_{i}\mid\hb)}
\end{eqnarray*}
where $\Phi(\cdot\mid\hb)$ is the normal cumulative distribution
function of $P(x_{i}\mid\hb)$. Now $P(x_{i}\mid\hb,\eb)$ is a truncated
normal distribution, from which we can sample using the simple rejection
method, or more advanced methods such as those in \cite{robert1995simulation}.

\subsection{Inference under Inequality Constraints}

For general \emph{inequality constraints} (Section~\ref{sub:Inequality-Constraints}),
the input variables are interdependent due to the linear transform
$A$. However, we can specify the conditional distribution $P(\x_{i}\mid\xb_{\neg i},\hb,\eb)$
(here $\xb_{\neg i}=\xb\backslash x_{i}$) by realising that\vspace{-0.1cm}

\[
b_{m}-\sum_{j\ne i}A_{mj}x_{j}\le A_{mi}x_{i}\le c_{m}-\sum_{j\ne i}A_{mj}x_{j}
\]
\vspace{-0.1cm}
 where $A_{mi}\ne0$ for $m=1,2,...,M$. In other words, $x_{i}$
is conditionally box-constrained given other variables.

This suggests a Gibbs procedure by looping through $x_{1},x_{2},...,x_{N}$.
With some abuse of notation, let $\tilde{b}_{mi}=\left(b_{m}-\sum_{j\ne i}A_{mj}x_{j}\right)/A_{mi}$
and $\tilde{c}_{mi}=\left(c_{m}-\sum_{j\ne i}A_{mj}x_{j}\right)/A_{mi}$.
The constraints can be summarised as 
\begin{eqnarray*}
x_{i} & \in & \cap_{m=1}^{M}\left[\min\left\{ \tilde{b}_{mi},\tilde{c}_{mi}\right\} ,\max\left\{ \tilde{b}_{mi},\tilde{c}_{mi}\right\} \right]\\
 & = & \left[\max_{m}\min\left\{ \tilde{b}_{mi},\tilde{c}_{mi}\right\} ,\min_{m}\max\left\{ \tilde{b}_{mi},\tilde{c}_{mi}\right\} \right]
\end{eqnarray*}
The $\min$ and $\max$ operators are needed to handle change in inequality
direction with the sign of $A_{mi}$, and the join operator is due
to multiple constraints.

For more sophisticated Gibbs procedures, we refer to the work in \cite{geweke1991efficient}.

\subsection{Estimating the Binary Posteriors \label{sub:Estimating-the-Posteriors}}

We are often interested in the posteriors $P(\hb\mid\eb)$, e.g.,
for further processing. Unlike the standard RBMs, the binary latent
variables here are coupled through the unknown Gaussians and thus
there are no exact solutions unless the evidences are all point assignments.
The MCMC-based techniques described above offer an approximate estimation
by averaging the samples $\left\{ \hb^{(s)}\right\} _{s=1}^{S}$.
For the case of boxed constraints, mean-field offers an alternative
approach which may be numerically faster. In particular, the mean-field
updates are recursive:\vspace{-0.1cm}

\begin{eqnarray*}
Q_{k} & \leftarrow & \frac{1}{1+\exp\left\{ -\gamma_{k}-\sum_{i}W_{ik}\hat{\mu}_{i}\right\} }\\
\mu_{i} & \leftarrow & \alpha_{i}+\sum_{k}W_{ik}Q_{k}\\
\hat{\mu}_{i} & = & \mu_{i}+\frac{\phi(b_{i}-\mu_{i})-\phi(c_{i}-\mu_{i})}{\Phi(c_{i}-\mu_{i})-\Phi(b_{i}-\mu_{i})}
\end{eqnarray*}
\vspace{-0.1cm}
 where $Q_{k}$ is the probability of the unit $k$ being activated,
$\hat{\mu}_{i}$ is the mean of the normal distribution truncated
in the interval $[b_{i},c_{i}]$, $\phi(z)$ is the probability density
function, and $\Phi(z)$ is normal cumulative distribution function.
Interested readers are referred to the Supplement%
\footnote{Supplement material will be available at: http://truyen.vietlabs.com%
} for more details.

\subsection{Estimating Probability of Evidence Generation \label{sub:Estimating-Probability-of-Evidences}}

Given the hidden states $\hb$ we want to estimate the probability
that hidden states generate a particular evidence $\eb$\vspace{-0.1cm}

\[
P(\eb\mid\hb)=\int_{\Domainb(\eb)}P(\xb\mid\hb)d\xb
\]
\vspace{-0.1cm}
 For boxed constraints, analytic solution is available since the Gaussian
variables are decoupled, i.e., $P(e_{i}\mid\hb)=\Phi(c_{i}-\mu_{i})-\Phi(b_{i}-\mu_{i})$,
where $\mu_{i}=\alpha_{i}+\sum_{k}W_{ik}h_{k}$. For general inequality
constraints, however, these variables are coupled by the inequalities.
The general strategy is to sample from $P(\xb\mid\hb)$ and compute
the portion of samples falling into the constrained domain $\Domainb(\eb)$.
For certain classes of inequalities we can approximate the Gaussian
by appropriate distributions from which the integration has the closed
form. In particular, those inequalities imposed by the categorical
and rank evidences can be dealt with by using the \emph{extreme value
distributions}. The integration will give the logit form on distribution
of categories and Plackett-Luce distribution of ranks. For details,
we refer to the Supplement.

\section{Stochastic Gradient Learning with Persistent Markov Chains \label{sub:Stochastic-Gradient-Learning}}

Learning is based on maximising the evidence likelihood\vspace{-0.1cm}

\begin{eqnarray*}
\LL & = & \log P(\eb)=\log\sum_{\hb}\int_{\Domainb(\eb)}P(\hb,\xb)d\xb
\end{eqnarray*}
\vspace{-0.1cm}
 where $P(\hb,\xb)$ is defined in Eq.~(\ref{eq:Model-def}). Let
$Z(\eb)=\sum_{\hb}\int_{\Domainb(\eb)}\exp\left\{ -E(\xb,\hb)\right\} d\xb$,
then $\LL=\log Z(\eb)-\log Z$. The gradient w.r.t. the mapping parameter
reads\vspace{-0.1cm}

\begin{eqnarray}
\partial_{W_{ik}}\LL & = & \mathbb{E}_{P(x_{i},h_{k}\mid\eb)}\left[x_{i}h_{k}\right]-\mathbb{E}_{P(x_{i},h_{k})}\left[x_{i}h_{k}\right]\label{eq:ll-grad}
\end{eqnarray}
\vspace{-0.1cm}
 The derivation is left to the Supplement.

\subsection{Estimating Data Statistics}

The \emph{data-dependent} statistics $\mathbb{E}_{P(x_{i},h_{k}\mid\eb)}\left[x_{i}h_{k}\right]$
and the \emph{data-independent} statistics\\
$\mathbb{E}_{P(x_{i},h_{k})}\left[x_{i}h_{k}\right]$ are not tractable
to compute in general, and thus approximations are needed.

\paragraph{Data-dependent statistics.}

Under the box constraints, the mean-field technique (Section~\ref{sub:Estimating-the-Posteriors})
can be employed as follows 
\[
\mathbb{E}_{P(x_{i},h_{k}\mid\eb)}\left[x_{i}h_{k}\right]\approx\hat{\mu}_{i}Q_{k}
\]
For general cases, sampling methods are applicable. In particular,
we maintain one persistent Markov chain \cite{younes1989parametric,tieleman2008training}
per data instance and estimate the statistics after a very short run.
This would explore the space of the data-dependent distribution $P(\xb,\hb\mid\eb)$
by alternating between $P(\hb\mid\xb)$ and $P(\xb\mid\hb,\eb)$ using
techniques described in Section~\ref{sub:Generalised-Inference}.

\paragraph{Data-independent statistics.}

Mean-field distributions are not appropriate for exploring the entire
state space because they tend to fit into one mode. One practical
solution is based on the idea of Hinton's Contrastive Divergence (CD),
where we create another Markov chain on-the-fly starting from the
latest state of the clamped chain. This chain will be discarded after
each parameter update. This is particular useful when the models are
instance-specific, e.g., in collaborative filtering, it is much cheaper
to build one model per user, all share the same parameters. If it
is not the case, then we can maintain a moderate set of parallel chains
and collect the samples after a short run at every updating step \cite{younes1989parametric,tieleman2008training}.

\subsection{Learning the Box Boundaries}

In the case of boxed constraints, sometimes it is helpful to learn
the boundaries $\left\langle b_{i},c_{i}\right\rangle $ themselves.
The gradient of the log-likelihood w.r.t. the lower boundaries reads{\small 
\begin{eqnarray*}
\partial_{b_{i}}\LL & = & \frac{1}{Z(\eb)}\sum_{\hb}\partial_{b_{i}}\int_{\Domainb(\eb)}\exp\left\{ -E(\xb,\hb)\right\} d\xb\\
 & = & \sum_{\hb}\partial_{b_{i}}\int_{\Domainb(\eb)}P\left(\xb,\hb\mid\eb\right)d\xb\\
 & \approx & \frac{1}{S}\partial_{b_{i}}\int_{b_{i}}^{c_{i}}\int_{\Domainb(\eb)\backslash[b_{i},c_{i}]}P\left(\xb\mid\hb^{(s)},\eb\right)d\xb_{\neg i}dx_{i}\\
 & = & -\frac{1}{S}\sum_{\hb}P\left(x_{i}=b_{i}\mid\hb^{(s)},\eb\right)
\end{eqnarray*}
}where $\left\{ \hb^{(s)}\right\} _{s=1}^{S}$ are samples collected
during the MCMC procedure running on the data-dependent distribution
$P\left(\xb,\hb\mid\eb\right)$. Similarly we would have the gradient
w.r.t. the upper boundaries: 
\[
\partial_{c_{i}}\LL=\frac{1}{S}\sum_{\hb}P\left(\xb=c_{i}\mid\hb^{(s)},\eb\right).
\]

\section{Applications \label{sec:Experiments}}

In this section, we describe applications of the $\Model$ for three
realistic domains, namely handwritten digit recognition, collaborative
filtering and worldwide survey analysis. Before going to the details,
let us first address key implementation issues (see Supplement for
more details).

One observed difficulty in training the $\Model$ is that the hidden
samples can get stuck in one of the two ends and thus learning cannot
progress. The reasons might be the large mapping parameters or the
unbounded nature of the underlying Gaussian variables, which can saturate
the hidden units. We can control the norm of the mapping parameters,
either by using the standard $\ell_{2}$-norm regularisation, or by
rescaling the norm of the parameter vector for each hidden unit. To
deal with the non-boundedness of the Gaussian variables, then we can
restrict their range, making the model bounded.

Another effective solution is to impose a constraint on the posteriors
by adding the regularising term to the log-likelihood, e.g., 
\[
\lambda\left\{ \sum_{k}\left[\rho\log P(h_{k}^{1}\mid\eb)+(1-\rho)\log\left(1-P(h_{k}^{1}\mid\eb)\right)\right]\right\} 
\]
where $\rho\in(0,1)$ is the expected probability that a hidden unit
will turn on given the evidence and $\lambda>0$ is the regularisation
weight. Maximising this quantity is essentially minimising the Kullback-Leibler
divergence between the expected posteriors and the true posteriors.
In our experiments, we found $\rho\in(0.1,0.3)$ and $\lambda\in(0.1,1)$
gave satisfying results.

The main technical issue is that $P(h_{k}^{1}\mid\eb)$ does not have
a simple form due to the integration over all the constrained Gaussian
variables. Approximation is thus needed. The use of mean-field methods
will lead to the simple sigmoid form, but it is only applicable for
boxed constraints since it breaks down deterministic constraints among
variables (Section~\ref{sub:Estimating-the-Posteriors}). However,
we can estimate the ``mean'' truncated Gaussian $\hat{\mu}_{i}$
by averaging the recent samples of the Gaussian variables in the data-dependent
phase.

Once these safeguards are in place, learning can greatly benefit from
quite large learning rate and small batches as it appears to quickly
get the samples out off the local energy traps by significantly distorting
the energy landscape. Depending on the problem sizes, we vary the
batch sizes in the range $[100,1000]$.

\subsection{Probit RBM for Handwritten Digits \label{sub:Probit-RBM-for-Handwritten-Digits}}

We use the name Probit RBM to denote the special case of $\Model$
where the observations are binary (i.e., boxed constraints, see Section~\ref{sub:Boxed-Constraints}).
The threshold $\theta_{i}$ for each visible unit $i$ is chosen so
that under the zero mean, the probability of generating a binary evidence
equals the empirical probability, i.e., $1-\Phi(\theta_{i})=\bar{e}_{i}$,
and thus $\theta_{i}=\Phi^{-1}(1-\bar{e}_{i})$. Since any mismatch
in thresholds can be corrected by shifting the corresponding biases,
we do not need to update the thresholds further.

\begin{figure}
\begin{centering}
\includegraphics[height=0.45\columnwidth]{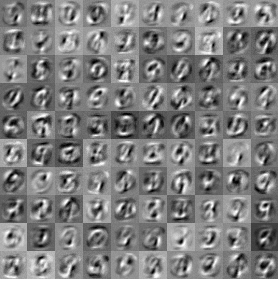}$\quad\quad$\includegraphics[width=0.45\columnwidth]{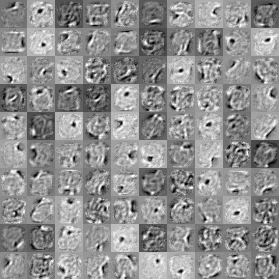} 
\par\end{centering}

\caption{$100$ MNIST feature weights (one image per hidden unit) learned by
Probit RBM (left) and RBM with CD-1 (right).\label{fig:Learned-MNIST-features}\vspace{-0.2cm}
 }
\end{figure}

We report here the result of the mean-field method for computing data-dependent
statistics, which are averaged over a random batch of $500$ images.
For the data-independent statistics, $500$ persistent chains are
run in parallel with samples collected after every $5$ Gibbs steps.
The sparsity level $\rho$ is set to $0.3$ and the sparseness weight
$\lambda$ is set to $0.5$. Once the model has been learned, mean-field
is used to estimate the hidden posteriors. Typically this mean-field
is quite fast as it converges in a few steps.

\begin{figure}
\begin{centering}
\includegraphics[width=0.75\columnwidth,height=0.7\columnwidth]{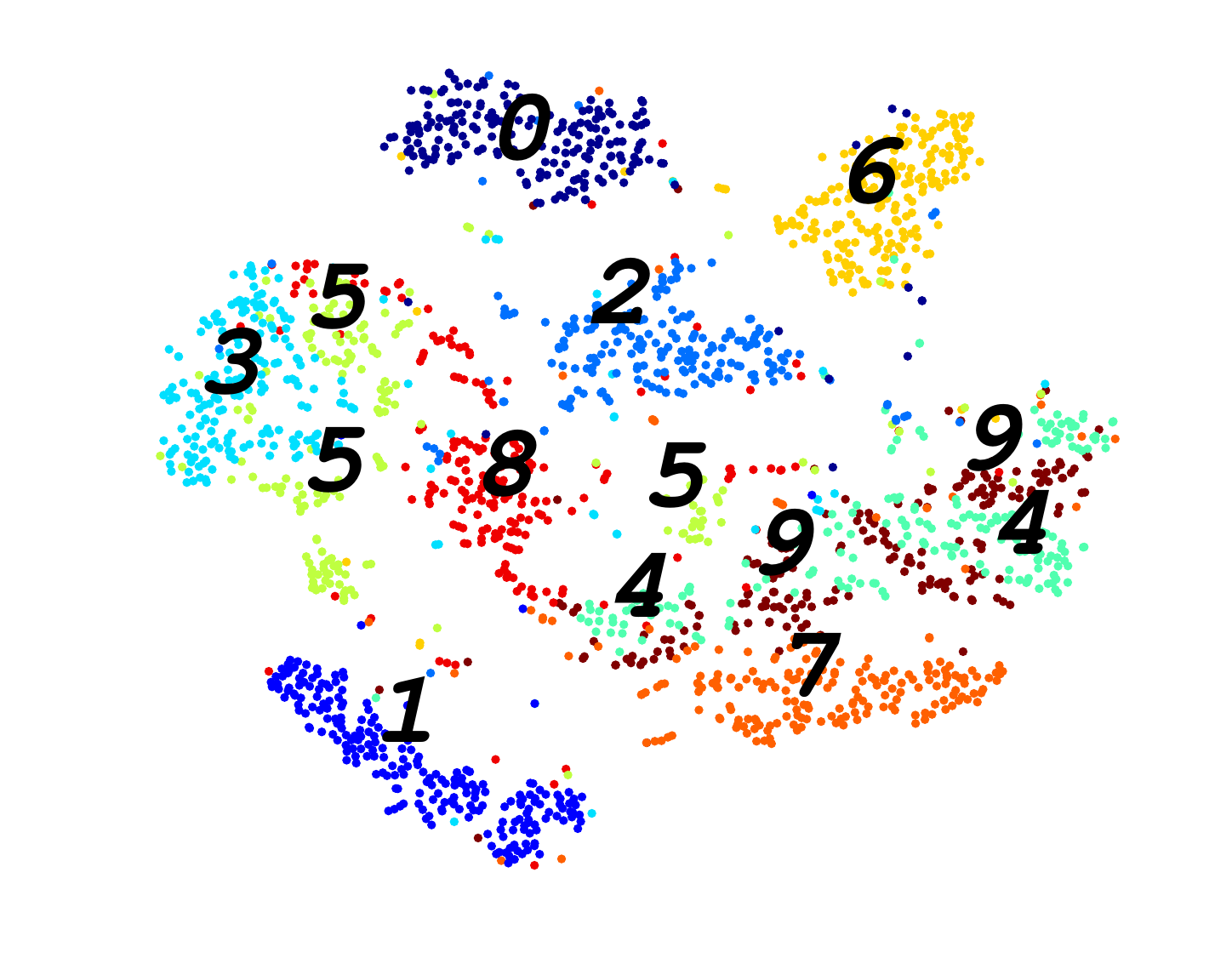}~~\includegraphics[width=0.17\columnwidth,height=0.65\columnwidth]{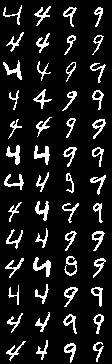} 
\par\end{centering}

\caption{t-SNE visualisation of the learned representations on MNIST (left)
and random samples of two confused digits (4,9) (right). Best viewed
in colours. \label{fig:t-SNE-visualisation-of-MNIST}\vspace{-0.2cm}
 }
\end{figure}

\begin{figure}
\begin{centering}
\includegraphics[width=0.45\columnwidth]{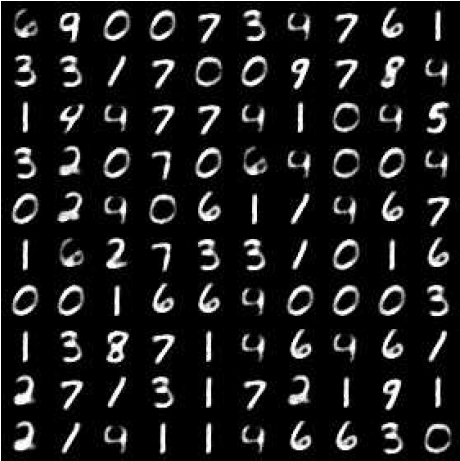}$\quad\quad$\includegraphics[width=0.45\columnwidth]{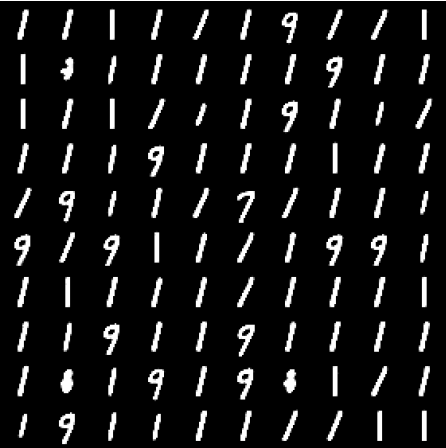} 
\par\end{centering}

\caption{Samples generated from the Probit RBM (left) and RBM with CD-1 (right)
by mode seeking procedures (see Supplement).\label{fig:Samples-generated-from-Probit-RBM}\vspace{-0.2cm}
 }
\end{figure}

We take the data from MNIST and binarize the images using a mid-intensity
threshold. The learned representation is shown in Figure~\ref{fig:t-SNE-visualisation-of-MNIST}.
Most digits are well separated in 2D except for digits $4$ and $9$.
The learned representation can be used for classifications, e.g.,
by feeding to the multiclass logistic classifier. For $500$ hidden
units, the Probit RBM achieves the error rate of $3.28$\%, comparable
with those obtained by the RBM trained with CD-$1$ ($3.02$\%), and
much better than the raw pixels ($8.46$\%). The features discovered
by the Probit RBM and RBM with CD-$1$ are very different (Figure~\ref{fig:Learned-MNIST-features}),
and this is expected because they operate on different input representations.
The energy surface learned by the Probit RBM is smooth enough to allow
efficient exploration of modes, as shown in Figure~\ref{fig:Samples-generated-from-Probit-RBM}.

\subsection{Rank Evidences for Collaborative Filtering}

In collaborative filtering, one of the main goals is to produce a
personalized ranked list of items. Until very recently, the majority
in the area, on the other hand, focused on predicting the ratings,
which are then used for ranking items. It can be arguably more efficient
to learn a rank model directly instead of going through the intermediate
steps.

We build one $\Model$ for ranking with ties (i.e., inequality constraints,
see Section~\ref{sub:Inequality-Constraints}) per user due to the
variation in item choices but all the $\Model$s share the same parameter
set. The handling of ties is necessary because during training, many
items share the same rating. Unseen items are simply not accounted
for in each model: We only need to compare the utilities between the
items seen by each person. The result is that the models are very
sparse and fast to run. For the data-dependent statistics, we maintain
one Markov chain per user. Since there is no single model for all
data instances, the data-independent statistics cannot be estimated
from a small set of Markov chains. Rather we also maintain a data-independent
chain per data instance, which can be persistent on their own, or
restarted from the data-dependent chains after every parameter updating
step. The latter case, which is reported here, is in the spirit of
the Hinton's Contrastive Divergence, where the data-independent chain
is just a few steps away from the data-dependent chain.

Once the model has been trained, the hidden posterior vector $\hat{\hb}=\left(\hat{h}_{1},\hat{h}_{2},...,\hat{h}_{K}\right)$,
where $\hat{h}_{k}=P(h_{k}=1\mid\eb_{u})$, is used as the new representation
of the tastes of user $u$. The rank of unseen movies is the mode
of the distribution $P(\eb^{*}\mid\hat{\hb})$, where $\eb^{*}$ are
the rank-based evidences (see Section~\ref{sub:Estimating-Probability-of-Evidences}).
For fast computation of $P(\eb^{*}\mid\hat{\hb})$, we approximate
the Gaussian by a Gumbel distribution, which leads to a simple way
of ranking movies using the mean ``utility'' $\mu_{ui}=\alpha_{i}+W_{i\bullet}\hat{\hb}_{u}$
for user $u$ (see the Supplement for more details).

The data used in this experiment is the MovieLens, which contains
$1$M ratings by approximately $6$K users on $4$K movies. To encourage
diversity in the rank lists, we remove the top $10$\% most popular
movies. We then remove users with less than $30$ ratings on the remaining
movies. The most recently rated $10$ movies per user are held out
for testing, the next most recent $5$ movies are used for tuning
hyper-parameters, and the rest for training.

For comparison, we implement a simple baseline using item popularity
for ranking, and thus offering a naive non-personalized solution.
For personalized alternatives, we implement two recent rank-based
matrix factorisation methods, namely ListRank.MF \cite{shi2010list}
and PMOP \cite{Truyen:2011a}. Two ranking metrics from the information
retrieval literature are used: the \emph{ERR }\cite{chapelle2009expected}
and the \emph{NDCG@T} \cite{jarvelin2002cumulated}. These metrics
place more emphasis on the top-ranked items. Table~\ref{tab:MovieLens-1M-rank}
reports the movie ranking results on test subset (each user is presented
with a ranked list of unseen movies), demonstrating that the $\Model$
is a clear winner in all metrics.

\begin{table}
\begin{centering}
\begin{tabular}{|r|cccc|}
\cline{2-5} 
\multicolumn{1}{r|}{} & \emph{ERR}  & \emph{N@1}  & \emph{N@5}  & \emph{N@10}\tabularnewline
\hline 
Item popularity  & 0.587  & 0.560  & 0.680  & 0.835\tabularnewline
\hline 
ListRank.MF  & 0.653  & 0.673  & 0.751  & 0.873\tabularnewline
\hline 
PMOP  & 0.648  & 0.664  & 0.747  & 0.871\tabularnewline
\hline 
\textbf{$\Model$}  & \textbf{0.678}  & \textbf{0.722}  & \textbf{0.792}  & \textbf{0.893}\tabularnewline
\hline 
\end{tabular}
\par\end{centering}

\caption{Item ranking results on MovieLens -- the higher the better ($K=50$).
Here \emph{N@T} is a shorthand for \emph{NDCG@T}. \label{tab:MovieLens-1M-rank}\vspace{-0.2cm}
 }
\end{table}

\subsection{Mixed Evidences for World Attitude Analysis \label{sub:Mixed-Evidences-for-GA}}

Finally, we demonstrate the $\Model$ on mixed evidences. The data
is from the survey analysis domain, which mostly consists of multiple
questions of different natures such as basic facts (e.g., ages and
genders) and opinions (e.g., binary choices, single choices, multiple
choices, ordinal judgments, preferences and ranks). The standard approach
to deal with such heterogeneity is to perform the so-called ``coding'',
which converts types into some numerical representations (e.g., ordinal
scales into stars, ranks into multiple pairwise comparisons) so that
standard processing tools can handle. However, this coding process
breaks the structure in the data and thus significant information
will be lost. Thus our $\Model$ offers a scalable and generic machinery
to process the data in its native format and then convert the mixed
types into a more homogeneous posterior vector.

We use the global attitude survey dataset collected by the PewResearch
Centre%
\footnote{The datasets are publicly available from http://pewresearch.org/%
}. The survey was conducted on $24,717$ people from $24$ countries
during the period of March 17 \textendash{} April 21, 2008 on a variety
of topics concerning people's life, opinions on issues in their countries
and around the world as well as future expectations. There are $52$
binary, $124$ categorical (of variable category sizes), $3$ continuous,
$165$ ordinal (of variable level sizes) question types.

Like the case of collaborative filtering, we build one $\Model$ per
respondent due to the variation in questions and answers but all the
$\Model$s share the same parameter set. Unanswered/inappropriate
questions are ignored. For each respondent, we maintain $2$ persistent
and non-interacting Markov chains for the data-dependent statistics
and the data-independent statistics, respectively.

\begin{figure}
\begin{centering}
\includegraphics[width=0.9\columnwidth,height=0.7\columnwidth]{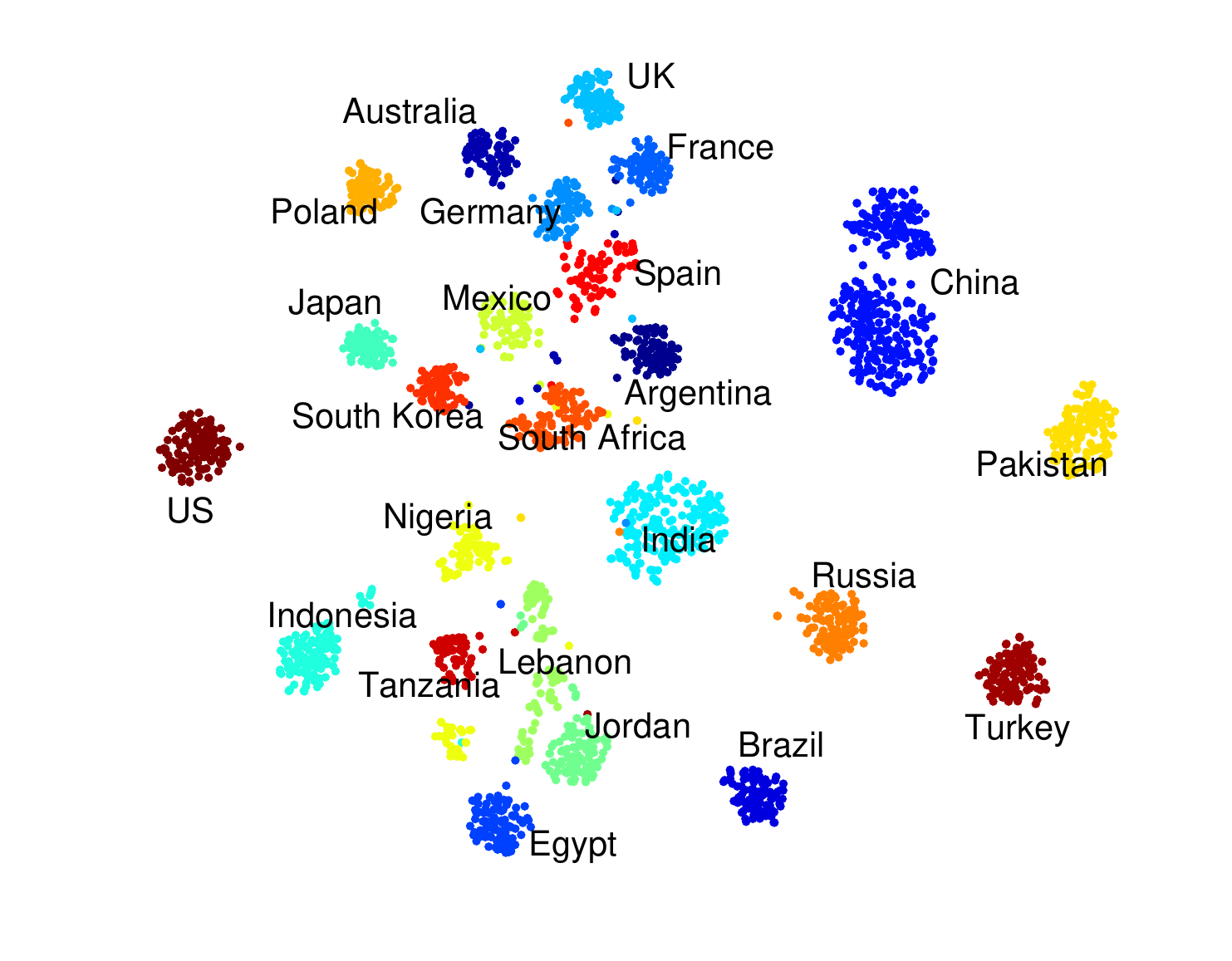} 
\par\end{centering}

\vspace{-0.2cm}

\caption{Distribution of global attitudes obtained by projecting the hidden
posteriors ($100$ hidden units) on 2D using t-SNE. A dot represents
one respondent. Best viewed in colours. \label{fig:Distribution-of-world-opinions}}

\vspace{-0.2cm}
\end{figure}

Figure~\ref{fig:Distribution-of-world-opinions} shows the 2D distribution
of respondents from $24$ countries obtained by feeding the posteriors
to the t-SNE \cite{van2008visualizing} (here no explicit information
of countries is used). It is interesting to see the cultural/social
clustering and gaps between countries as opposed to the geographical
distribution (e.g., between Indonesia and Egypt, Australia and UK
and the relative separation of the China, Pakistan, Turkey and the
US from the rest). To predict the $24$ countries, we feed the posteriors
into the standard multiclass logistic regression and achieve an error
rate of $0.49\%$, suggesting that the $\Model$ has captured the
intrastate regularities and separated the interstate variations well.\vspace{-0.2cm}

\section{Related Work \label{sec:Related-Work}}

Latent multivariate Gaussian variables have been widely studied in
statistical analysis, initially to model correlated binary data%
\footnote{This is often known as \emph{multivariate probit models}.%
} \cite{ashford1970multi,chib1998analysis} then now used for a variety
of data types such as ordered categories \cite{kottas2005nonparametric},
unordered categories \cite{zhang2008bayesian}, and the mixture of
types\emph{ }\cite{dunson2005bayesian}\emph{. }Learning with the
underlying Gaussian model is notoriously difficult for large-scale
setting: independent sampling costs cubic time due to the need of
inverting the covariance matrix, while MCMC techniques such as Gibbs
sampling can be very slow if the graph is dense and the interactions
between variables are strong. This can be partly overcome by adding
one more layer of latent variables as in factor analysis \cite{wedel2001factor,khan2010variational}
and probabilistic principle component analysis \cite{tipping1999probabilistic}.
The main difference from our $\Model$ is that those models are directed
with continuous factors while ours is undirected with binary factors.

Gaussian RBMs have been used for modelling continuous data such as
visual features \cite{hinton2006rdd}, where the evidences are the
value assignments, and thus a limiting case of our evidence system.
Some restrictions to the continuous Boltzmann machines have been studied:
In \cite{downs2000nonnegative}, Gaussian variables are assumed to
be non-negative, and in \cite{yasuda2007boltzmann}, continuous variables
are bounded. However, we do not make these restrictions on the model
but rather placing restrictions during the training phase only. GRBMs
that handle ordinal evidences have been studied in \cite{Truyen:2012d},
which is an instance of the boxed-constraints in our $\Model$.

\section{Discussion and Conclusion \label{sec:Conclusion}}

Since the underlying variables of the $\Model$ are Gaussian, various
extensions can be made without much difficulty. For example, direct
correlations among variables, \emph{regardless of their types}, can
be readily modelled by introducing the non-identity covariance matrix
\cite{ranzato2010modeling}. This is clearly a good choice for image
modelling since nearby pixels are strongly correlated. Another situation
is when the input units are associated with their own attributes.
Each unit can be extended naturally by adding a linear combination
of attributes to the mean structure of the Gaussian.

The additive nature of the mean-structure allows the natural extension
to matrix modelling (e.g., see \cite{Truyen:2009a,Truyen:2012d}).
That is, we do not distinguish the role of rows and columns, and thus
each row and column can be modelled using their own hidden units (the
row parameters and columns parameters are different). Conditioned
on the row-based hidden units, we return to the standard $\Model$
for column vectors. Inversely, conditioned on the column-based hidden
units, we have the $\Model$ for row vectors.

To sum up, we have proposed a generic class of models called Thurstonian
Boltzmann machine ($\Model$) to unify many type-specific modelling
problems and generalise them to the general problem of learning from
multiple groups of inequalities. Our framework utilises the Gaussian
restricted Boltzmann machines, but the Gaussian variables are never
observed except for one limiting case. Rather, those variables are
subject to inequality constraints whenever an evidence is observed.
Under this representation, the $\Model$ supports a very wide range
of evidences, many of which were not possible before in the Boltzmann
machine literature, \emph{without} the need to specify type-specific
models. In particular, the $\Model$ supports any combination of the
point assignments, intervals, censored values, binary, unordered categories,
multi-categories, ordered categories, (in)-complete ranks with and
without ties.

We demonstrated the $\Model$ on three applications of very different
natures, namely handwritten digit recognition, collaborative filtering
and complex survey analysis. The results are satisfying and the performance
is competitive with those obtained by type-specific models.

\appendix

\section{Supplementary Material}

\subsection{Inference}

\subsubsection{Estimating the Partition Function \label{sub:Estimating-the-Partition}}

For convenience, let us re-parameterise the distribution as follows
\begin{eqnarray}
\phi_{i}(\x_{i}) & = & \exp\left\{ -\frac{\x_{i}^{2}}{2}+\alpha_{i}\x_{i}\right\} \label{eq:local-variance}\\
\psi_{ik}(\x_{i},h_{k}) & = & \exp\left\{ W_{ik}x_{i}h_{k}\right\} \nonumber \\
\phi_{k}(h_{k}) & = & \exp\left\{ \gamma_{k}h_{k}\right\} \nonumber 
\end{eqnarray}

The model potential is then the product of all local potentials 
\begin{equation}
\Psi(\ub,\hb)=\left[\prod_{i}\phi_{i}(\x_{i})\right]\left[\prod_{ik}\psi_{ik}(\x_{i},h_{k})\right]\left[\prod_{k}\phi_{k}(h_{k})\right]\label{eq:potential-factorise}
\end{equation}
The partition function can be rewritten as 
\begin{eqnarray*}
Z & = & \sum_{\hb}\int_{\ub}\Psi(\ub,\hb)d\ub\\
 & = & \sum_{\hb}\Omega(\hb)
\end{eqnarray*}
where $\Omega(\hb)=\int_{\ub}\Psi(\ub,\hb)d\ub$. We now proceed to
compute $\Omega(\hb)$: 
\begin{eqnarray*}
\Omega(\hb) & = & \left[\prod_{k}\phi_{k}(h_{k})\right]\int_{\ub}\left[\prod_{i}\phi_{i}(\x_{i})\right]\left[\prod_{ik}\psi_{ik}(\x_{i},h_{k})\right]d\ub\\
 & = & \left[\prod_{k}\phi_{k}(h_{k})\right]\prod_{i}\int_{x_{i}}\exp\left\{ -\frac{\x_{i}^{2}}{2}+(\alpha_{i}+\sum_{k}W_{ik}h_{k})\x_{i}\right\} dx_{i}\\
 & = & \left[\prod_{k}\phi_{k}(h_{k})\right]\prod_{i}C_{i}\int_{x_{i}}\exp\left\{ -\frac{\left(x_{i}-\mu_{i}(\hb)\right)^{2}}{2}\right\} dx_{i}\\
 & = & \left[\prod_{k}\phi_{k}(h_{k})\right]\prod_{i}C_{i}\sqrt{2\pi\sigma_{i}^{2}}
\end{eqnarray*}
where

\begin{eqnarray*}
\mu_{i}(\hb) & = & \alpha_{id}+\sum_{k=1}^{K}W_{ik}h_{k}\\
C_{i} & = & \exp\left\{ \frac{1}{2}\left(\frac{\mu_{i}(\hb)}{\sigma_{i}}\right)^{2}\right\} 
\end{eqnarray*}

Now we can define the distribution over the hidden layer as follows
\[
P(\hb)=\frac{1}{Z}\Omega(\hb)
\]

Now we apply the Annealed Importance Sampling (AIS) \cite{neal2001annealed}.
The idea is to introduces the notion of inverse-temperature $\tau$
into the model, i.e., $P(\hb|\tau)\propto\Omega(\hb)^{\tau}$.

Let $\{\tau_{s}\}_{s=0}^{S}$ be the (slowly) increasing sequence
of temperature, where $\tau_{0}=0$ and $\tau_{S}=1$, that is $\tau_{0}<\tau_{1}...<\tau_{S}$.
At $\tau_{0}=0$, we have a uniform distribution, and at $\tau_{S}=1$,
we obtain the desired distribution. At each step $s$, we draw a sample
$\hb^{s}$ from the distribution $P(\hb|\tau_{s-1})$ (e.g. using
some Metropolis-Hastings procedure). Let $P^{*}(\hb|\tau)$ be the
unnormalised distribution of $P(\hb|\tau)$, that is $P(\hb|\tau)=P^{*}(\hb|\tau)/Z(\tau)$.
The final weight after the annealing process is computed as 
\begin{align*}
\omega & =\frac{P^{*}(\hb^{1}|\tau_{1})}{P^{*}(\hb^{1}|\tau_{0})}\frac{P^{*}(\hb^{2}|\tau_{2})}{P^{*}(\hb^{2}|\tau_{1})}...\frac{P^{*}(\hb^{S}|\tau_{S})}{P^{*}(\hb^{S}|\tau_{S-1})}
\end{align*}

The above procedure is repeated $T$ times. Finally, the normalisation
constant at $\tau=1$ is computed as $Z(1)\approx Z(0)\left(\sum_{t=1}^{T}\omega^{(t)}/T\right)$
where $Z(0)=2^{K}$, which is the number of configurations of the
hidden variables $\hb$.

\subsubsection{Estimating Posteriors using Mean-field \label{sub:Mean-field-Updates}}

Recall that for evidence $\eb$ we want to estimate posteriors $P(\hb\mid\eb)=\sum_{\xb\in\Omega(\eb)}P_{\Omega(\eb)}(\hb,\xb\mid\eb)$.
Assume that the evidences can be expressed in term of boxed constraints,
which lead to the following factorisation 
\[
P(\xb\mid\eb,\hb)=\prod_{i}P(x_{i}\mid\eb,\hb)
\]
This factorisation is critical because it ensures that there are no
deterministic constraints among $\left\{ x_{i}\right\} _{i=1}^{n}$,
which are the conditions that variational methods such as mean-fields
would work well. This is because mean-field solution will generally
not satisfy deterministic constraints, and thus may assign non-zeros
probability to improbably areas.

To be more concrete, the mean-field approximation would be $Q(\hb,\xb)\approx P(\hb,\xb\mid\eb)$
\begin{eqnarray*}
Q(\hb,\xb) & = & \prod_{k}Q_{k}(h_{k})\prod_{i}Q_{i}(x_{i})\\
\mbox{s.t.}\quad\xb & \in & \Omega(\eb)
\end{eqnarray*}
The best mean-field approximation will be the minimiser of the Kullback-Leibler
divergence 
\begin{eqnarray}
\mathcal{D}\left(Q||P\right) & = & \sum_{\hb}\sum_{\xb\in\Omega(\eb)}Q(\hb,\xb)\log\frac{Q(\hb,\xb)}{P(\hb,\xb\mid\eb)}\nonumber \\
 & = & -H\left[Q(\hb,\xb)\right]-\sum_{\hb}\sum_{\xb\in\Domainb(\eb)}Q(\hb,\xb)\log P(\hb,\xb\mid\eb)\label{eq:KL-mean-field}
\end{eqnarray}
where $H\left[Q(\hb,\xb)\right]$ is the entropy function. Now first,
exploit the fact that $Q$ is factorisable, and thus its entropy is
decomposable, i.e., 
\begin{equation}
H\left[Q(\hb,\xb)\right]=\sum_{k}H\left[Q_{k}(h_{k})\right]+\sum_{i}H\left[Q_{i}(x_{i})\right]\label{eq:entropy-mean-field-decompose}
\end{equation}

Second recall from Eq.~(\ref{eq:model-cond-prob}) that 
\[
P(\hb,\xb\mid\eb)=\frac{1}{Z(\eb)}\exp\left\{ -E(\xb,\hb)\right\} 
\]
and thus 
\[
\sum_{\hb}\sum_{\xb\in\Domainb(\eb)}Q(\hb,\xb)\log P(\hb,\xb\mid\eb)=-\sum_{\hb}\sum_{\xb\in\Domainb(\eb)}Q(\hb,\xb)E(\xb,\hb)-\log Z(\eb)
\]
Since $\log Z(\eb)$ is a constraint w.r.t. $Q(\hb,\xb)$, we can
safely ignore it here.

Now since $E(\xb,\hb)$ is decomposable (see Eq.~(\ref{eq:energy-def})),
we have 
\begin{eqnarray*}
\sum_{\hb}\sum_{\xb\in\Domainb(\eb)}Q(\hb,\xb)E(\xb,\hb) & = & \left(\sum_{i}\sum_{x_{i}\in\Domain(e_{i})}Q_{i}(x_{i})E_{i}(x_{i})\right)+\left(\sum_{k}\sum_{h_{k}}Q_{k}(h_{k})E_{k}(h_{k})\right)+\\
 &  & \quad+\left(\sum_{i}\sum_{k}\sum_{x_{i}\in\Domain(e_{i})}\sum_{h_{k}}Q_{i}(x_{i})Q_{k}(h_{k})E_{ik}(x_{i},h_{k})\right)
\end{eqnarray*}
where 
\begin{eqnarray*}
E_{i}(x_{i}) & = & \frac{\x_{i}^{2}}{2}-\alpha_{i}x_{i}\\
E_{k}(h_{k}) & = & -\gamma_{k}h_{k}\\
E_{ik}(x_{i},h_{k}) & = & -W_{ik}x_{i}h_{k}
\end{eqnarray*}

Combining this decomposition and Eq.~(\ref{eq:entropy-mean-field-decompose}),
we have completely decomposed the Kullback-Leibler divergence in Eq.~(\ref{eq:KL-mean-field})
into local terms:{\small 
\begin{eqnarray*}
\mathcal{D}\left(Q||P\right) & = & \sum_{i}\sum_{x_{i}\in\Domain(e_{i})}Q_{i}(x_{i})E_{i}(x_{i})+\sum_{k}\sum_{h_{k}}Q_{k}(h_{k})E_{k}(h_{k})+\\
 &  & +\left(\sum_{i}\sum_{k}\sum_{x_{i}\in\Domain(e_{i})}\sum_{h_{k}}Q_{i}(x_{i})Q_{k}(h_{k})E_{ik}(x_{i},h_{k})\right)-\sum_{i}H\left[Q_{i}(x_{i})\right]-\sum_{k}H\left[Q_{k}(h_{k})\right]
\end{eqnarray*}
}{\small \par}

Now we wish to minimise the divergence with respect to the local distributions
$\left\{ Q_{i}(x_{i}),Q_{k}(h_{k})\right\} $ for $i=1,2,...,N$ and$k=1,2,...,K$
knowing the proper distribution constraints 
\begin{eqnarray*}
\int_{x_{i}\in\Domain(e_{i})}Q_{i}(x_{i}) & = & 1\\
\sum_{h_{k}}Q_{k}(h_{k}) & = & 1
\end{eqnarray*}

By the method of Lagrangian multiplier, we have 
\[
L(\lambda)=\mathcal{D}\left(Q||P\right)+\sum_{i}\lambda_{i}\left(\int_{x_{i}\in\Domain(e_{i})}Q_{i}(x_{i})-1\right)+\sum_{k}\kappa_{k}\left(\sum_{h_{k}}Q_{k}(h_{k})-1\right)
\]

\begin{itemize}
\item Let us compute the partial derivative w.r.t. $Q_{i}(x_{i})$: 
\begin{eqnarray*}
\partial_{Q_{i}(x_{i})}L(\lambda) & = & \log Q_{i}(x_{i})+1+E_{i}(x_{i})+\sum_{k}Q_{k}(h_{k}^{1})E_{ik}(x_{i},h_{k}^{1})+\lambda_{i}
\end{eqnarray*}
where $h_{k}^{1}$ is a short hand for $h_{k}=1$ and we have made
use of the fact that $E_{ik}(x_{i},h_{k}=0)=0$. Setting this gradient
to zero yields 
\begin{eqnarray}
Q_{i}(x_{i}) & = & \exp\left\{ -\left(E_{i}(x_{i})+\sum_{k}Q_{k}(h_{k}^{1})E_{ik}(x_{i},h_{k}^{1})\right)-1-\lambda_{i}\right\} \nonumber \\
 & = & \exp\left\{ -\frac{1}{2}\left(x_{i}-\left(\alpha_{i}+\sum_{k}Q_{k}(h_{k}^{1})W_{ik}\right)\right)^{2}-1-\lambda_{i}\right\} \label{eq:mean-field-normal}
\end{eqnarray}
for $x_{i}\in\Domain(e_{i})$. Normalising this distribution would
lead to the truncated form of the normal distribution those the mean
is 
\begin{equation}
\mu_{i}=\alpha_{i}+\sum_{k}Q_{k}(h_{k}^{1})W_{ik}h_{k}\label{eq:mean-field-mean}
\end{equation}

\item In a similar way, the partial derivative w.r.t. $Q_{k}(h_{k})$ would
be 
\[
\partial_{Q_{k}(h_{k})}L(\lambda)=\log Q_{k}(h_{k})-h_{k}\left(\gamma_{k}+\sum_{i}W_{ik}\sum_{x_{i}\in\Domain(e_{i})}Q_{i}(x_{i})x_{i}\right)+1+\kappa_{k}
\]
Equating the gradient to zero, we have 
\[
Q_{k}(h_{k})\propto\exp\left\{ h_{k}\left(\gamma_{k}+\sum_{i}W_{ik}\hat{\mu}_{i}\right)\right\} 
\]
where $\hat{\mu}_{i}$ is the mean of the truncated normal distribution
\[
\hat{\mu}_{i}=\sum_{x_{i}\in\Domain(e_{i})}Q_{i}(x_{i})x_{i}
\]
Normalising $Q_{k}(h_{k})$ would lead to 
\begin{equation}
Q_{k}(h_{k}^{1})=\left[1+\exp\left\{ -\gamma_{k}-\sum_{i}W_{ik}\hat{\mu}_{i}\right\} \right]^{-1}\label{eq:mean-field-sigmoid}
\end{equation}

\item Finally, combining these findings in Eqs.~(\ref{eq:mean-field-normal},\ref{eq:mean-field-mean},\ref{eq:mean-field-sigmoid}),
and letting $\Domain(e_{i})=[b_{i},c_{i}]$ be the boxed constraint,
and using the fact that the mean of the truncated distribution is
\[
\hat{\mu}_{i}=\mu_{i}+\frac{\phi(b_{i}-\mu_{i})-\phi(c_{i}-\mu_{i})}{\Phi(c_{i}-\mu_{i})-\Phi(b_{i}-\mu_{i})}
\]
we would arrive at the three recursive equations 
\begin{eqnarray*}
q_{k} & \leftarrow & \frac{1}{1+\exp\left\{ -\gamma_{k}-\sum_{i}W_{ik}\hat{\mu}_{i}\right\} }\\
\mu_{i} & \leftarrow & \alpha_{i}+\sum_{k}W_{ik}q_{k}\\
\hat{\mu}_{i} & \leftarrow & \mu_{i}+\frac{\phi(b_{i}-\mu_{i})-\phi(c_{i}-\mu_{i})}{\Phi(c_{i}-\mu_{i})-\Phi(b_{i}-\mu_{i})}
\end{eqnarray*}
where $q_{k}$ is a short hand for $Q_{k}(h_{k}^{1})$ and$\phi(z)$
is the normal probability density function, and $\Phi(z)$ is the
cumulative distribution function $\clubsuit$ 
\end{itemize}

\subsubsection{Seeking Modes and Generating Representative Samples}

Once the model has been learned, samples can be generated straightforwardly
by first sampling the underlying Gaussian RBM and then collect the
true samples that satisfy the inequalities of interest. For example,
for binary samples, if the generated Gaussian value for a visible
unit is larger than the threshold, then we have an active sample.
Likewise, rank samples, we only need to rank the sampled Gaussian
values.

However, this may suffer from the poor mixing if we use standard Gibbs
sampling, that is the Markov chain may get stuck in some energy traps.
To jump out of the trap we propose to periodically raise the temperature
to a certain level (e.g., $10$) and then slowly cool down to the
original temperature (which is $1$). In our experiment, the cooling
is scheduled as follows 
\[
T\leftarrow\eta T
\]
where $\eta\in(0,1)$ is estimated so that for $n$ steps, the temperature
will drop from $T_{max}$ to $T_{min}$. That is, $T_{min}=\eta^{n}T_{max}$,
leading to $\eta=\left(T_{min}/T_{max}\right)^{1/n}$.

To locate a basis of attraction, we can lower the temperature further
(e.g., to $0.1$) to trap the particles there. Then we collect $k$
successive samples and take the average to be the representative sample.
In our experiments, $k=50$.

\subsection{Learning}

\subsubsection{Gradient of the Likelihood}

The log-likelihood of an evidence is 
\begin{eqnarray*}
\LL & = & \log P(\eb)=\log\sum_{\hb}\int_{\Domainb(\eb)}P(\hb,\xb)d\xb\\
 & = & \log\sum_{\hb}\int_{\Domainb(\eb)}\exp\left\{ -E(\xb,\hb)\right\} d\xb-\log Z\\
 & = & \log Z(\eb)-\log Z
\end{eqnarray*}
where $Z(\eb)=\sum_{\hb}\int_{\Domainb(\eb)}\exp\left\{ -E(\xb,\hb)\right\} d\xb$.
The gradient of $\log Z(\eb)$ w.r.t. the mapping parameter $W_{ik}$
reads 
\begin{eqnarray}
\partial_{W_{ik}}\log Z(\eb) & = & \frac{-1}{Z(\eb)}\sum_{\hb}\int_{\Domainb(\eb)}\exp\left\{ -E(\xb,\hb)\right\} \partial_{W_{ik}}E(\xb,\hb)d\xb\nonumber \\
 & = & -\sum_{\hb}\int_{\Domainb(\eb)}P(\xb,\hb\mid\eb)\partial_{W_{ik}}E(\xb,\hb)d\xb\label{eq:gradient-of-empi-Z}
\end{eqnarray}
where we have moved the constant $Z^{-1}(\eb)$ into the sum and integration
and make use of the fact that 
\begin{equation}
P(\xb,\hb\mid\eb)=\frac{P(\xb,\hb)}{P(\eb)}=\frac{1}{Z(\eb)}\exp\left\{ -E(\xb,\hb)\right\} \label{eq:model-cond-prob}
\end{equation}
where the domain of the Gaussian is constrained to $\xb\in\Domainb(\eb)$.

From the definition of the energy function in Eq.~(\ref{eq:energy-def}),
we know that the energy is decomposable, and thus the gradient w.r.t.
$W_{ik}$ only involves the pair $(x_{i},h_{k})$. In particular 
\[
\partial_{W_{ik}}E(\xb,\hb)=-x_{i}h_{k}
\]

This simplifies Eq.~(\ref{eq:gradient-of-empi-Z}) 
\begin{eqnarray*}
\partial_{W_{ik}}\log Z(\eb) & = & -\sum_{h_{i}}\int_{\Domain(e_{i})}P(x_{i},h_{k}\mid\eb)\partial_{W_{ik}}E(\xb,\hb)dx_{i}\\
 & = & \mathbb{E}_{P(\xb,\hb\mid\eb)}\left[x_{i}h_{k}\right]
\end{eqnarray*}

A similar process would lead to 
\[
\partial_{W_{ik}}\log Z=\mathbb{E}_{P(x_{i},h_{k})}\left[x_{i}h_{k}\right]
\]
and finally: 
\begin{eqnarray*}
\partial_{W_{ik}}\LL & = & \mathbb{E}_{P(x_{i},h_{k}\mid\eb)}\left[x_{i}h_{k}\right]-\mathbb{E}_{P(x_{i},h_{k})}\left[x_{i}h_{k}\right]\clubsuit
\end{eqnarray*}

\subsubsection{Regularising the Markov Chains}

One undesirable feature of the MCMC chains used in learning we have
experiences so far is the tendency for the binary hidden states to
get stuck, i.e., after some point they do not flip their assignments
as learning progresses. We conjecture that this phenomenon may be
due to the \emph{saturation effect} inherent in the factor posterior:
\[
P(h_{k}=1\mid\ub)=\frac{1}{1+\exp\left(-\gamma_{k}-\sum_{i}W_{ik}x_{i}\right)}
\]
i.e., once the collected value to a node $\left(\gamma_{k}+\sum_{i}W_{ik}x_{i}\right)$
is too high or too low, it is very hard to over turn.

Fortunately, there is a known technique to regularise the chain: we
enforce that at a time, there should be only a fraction $\rho$ of
nodes which are active, where $\rho\in(0,1)$. One way is to maximise
the following objective function 
\begin{eqnarray*}
\LL_{2} & = & \LL+\lambda\int\left[\sum_{k}\sum_{h_{k}}\rho(h_{k})\log P(h_{k}\mid\ub)\right]P(\ub\mid\vb)d\ub
\end{eqnarray*}
where $\lambda>0$ is the weighting factor, $\rho(h_{k})=\rho$ if
$h_{k}=1$ and $\rho(h_{k})=1-\rho$ otherwise.

The gradient with respect to $g_{k}=\left(\gamma_{k}+\sum_{i}W_{ik}x_{i}\right)$
is then

\begin{eqnarray*}
\partial_{g_{k}}\LL_{2} & = & \partial_{g_{k}}\LL+\lambda\int\left[\partial_{g_{k}}\sum_{h_{k}}\rho(h_{k})\log P(h_{k}\mid\ub)\right]P(\ub|\vb)d\ub\\
 & = & \partial_{g_{k}}\LL+\lambda\int\left[\rho-P(h_{k}^{1}\mid\ub)\right]P(\ub|\vb)d\ub\\
 & \approx & \partial_{g_{k}}\LL+\frac{1}{S}\lambda\sum_{s}\left[\rho-P(h_{k}^{1}\mid\ub^{(s)})\right]
\end{eqnarray*}
where $S$ is the number of samples and $P(h_{k}^{1}\mid\ub)$ is
a shorthand for $P(h_{k}=1\mid\ub)$. Using the chain rule, we have:

\begin{eqnarray*}
\partial_{\gamma_{k}}\LL_{2} & \approx & \partial_{\gamma_{k}}\LL+\frac{1}{S}\lambda\sum_{s}\left[\rho-P(h_{k}^{1}\mid\ub^{(s)})\right]\\
\partial_{W_{ik}}\LL_{2} & \approx & \partial_{w_{ikd}}\LL+\frac{1}{S}\lambda\sum_{s}x_{i}\left[\rho-P(h_{k}^{1}\mid\ub^{(s)})\right]
\end{eqnarray*}

\subsubsection{Online Estimation of Posteriors \label{sub:Online-Estimation-of-Posteriors}}

For tasks such as data completion (e.g., collaborative filtering)
we need the posteriors $P(\hb\mid\vb)$ for the prediction phase.
One way is to run the Markov chain or doing mean-field from scratch.
Here we suggest a simple way to obtain an approximation directly from
the training phase without any further cost. The idea is to update
the estimated posterior $\hat{\hb}$ at each learning step $t$ in
an \emph{exponential smoothing} fashion:

\[
\hat{\hb}^{(t)}\leftarrow\eta\hat{\hb}^{(t-1)}+(1-\eta)\hat{\hb}^{(t)}
\]
for some \emph{smoothing factor} $\eta\in(0,1)$ and initial $\hat{\hb}^{(0)}$,
where $\bar{h}_{k}^{(t)}=P(h_{k}^{1}\mid\ub^{(t)},\vb)$ and $\ub^{(t)}$
is the sampled Gaussian at time $t$.

As learning progresses, early samples, which are from incorrect models,
will be exponentially weighted down. Typically we choose $\eta$ close
to $1$, e.g., $\eta=0.9$.

\subsubsection{Monitoring the Learning Progress}

It is often of practical importance to track the learning progress,
either by the reconstruction errors or by the data likelihood. The
data likelihood can be estimated as 
\[
P(\eb)\approx\frac{1}{S}\sum_{s=1}^{S}\int_{\Domainb(\eb)}P(\xb\mid\hb^{(s)})d\xb
\]
where $\hb^{(s)}$ are those samples collected as learning progressed
in the data-independent phase, and the integration can be carried
out using the technique described in the main text.

\subsection{Extreme Value Distributions}

Extreme value distributions are a class of distributions of extremal
measurements \cite{gumbel1958statistical}. Here we are concerned
about the popular Gumbel's distribution.

\subsubsection{Gumbel Distribution for Categorical Choices \label{sub:Gumbel-Distribution-Cat}}

Let us start from the Gumbel density function 
\[
P(x)=\frac{1}{\sigma}\exp\left\{ -\left(\frac{x-\mu}{\sigma}+e^{-\frac{x-\mu}{\sigma}}\right)\right\} 
\]
where $\mu$ is the mode (location) and $\sigma$ is the scale parameter.

Using Laplace's approximation (e.g., via Taylor's expansion of $\left(\frac{x-\mu}{\sigma}+e^{-\frac{x-\mu}{\sigma}}\right)$
using the second-order polynomial around $\mu$), we have 
\[
P(x)\propto\frac{1}{e\sigma}\exp\left\{ -\frac{(x-\mu)^{2}}{2\sigma^{2}}\right\} 
\]
Renormalising this distribution, e.g., by replacing $e\approx2.7183$
by $\sqrt{2\pi}\approx2.5066$ we obtain the standard Gaussian distribution.
Thus, \emph{we can use the Gumbel as an approximation to the Gaussian
distribution}.

Now we turn to the categorical model using Gumbel variables. We maintain
one variable per category, which plays the role of the utility for
the category. Assume that all utilities share the same scale parameter
$\sigma$. The existing literature \cite{mcfadden1973conditional}
asserts that the probability of choosing the $m$-th category is 
\[
P(e=c_{m})=\frac{e^{\mu_{m}/\sigma}}{\sum_{l}e^{\mu_{l}/\sigma}}
\]
where $\mu_{l}$ is the location of the $l$-th utility.

When we choose a the $m$-th category we must ensure that $x_{m}>\max_{l\ne m}x_{l}$.

Let $y_{l}=\exp\left(-\frac{x_{l}-\mu_{l}}{\sigma}\right)$ or equivalently
$x_{l}=\mu_{l}-\sigma\log y_{l}$. Thus $x_{l}<x_{m}$ means $\mu_{l}-\sigma\log y_{l}<\mu_{m}-\sigma\log y_{m}$,
or $y_{l}>y_{m}\exp\left(\frac{\mu_{l}-\mu_{m}}{\sigma}\right)$.

The CDF of the $l$-th Gumbel distribution is 
\begin{eqnarray*}
F_{l}(x_{m}) & = & \exp\left(-e^{-\frac{x_{m}-\mu_{l}}{\sigma}}\right)\\
 & = & \exp\left(-e^{-\frac{x_{m}-\mu_{m}}{\sigma}}e^{\frac{\mu_{l}-\mu_{m}}{\sigma}}\right)\\
 & = & \exp\left(-y_{m}e^{\frac{\mu_{l}-\mu_{m}}{\sigma}}\right)
\end{eqnarray*}

Thus choosing category $c_{m}$ would mean 
\begin{eqnarray*}
P(e=c_{m}) & = & \int P(x_{m})\left(\prod_{l\ne m}\int^{xm}P(x_{l})dx_{l}\right)dx_{m}\\
 & = & \int P(x_{m})\prod_{l\ne m}F_{l}(x_{m})dx_{m}
\end{eqnarray*}

We rewrite the Gumbel density function by changing variable from $x_{m}$
to $y_{m}$: 
\[
P(y_{m})=\frac{1}{\sigma}y_{m}\exp\left\{ -y_{m}\right\} 
\]
for $y_{m}\ge0$. Thus 
\[
P(x_{m})\prod_{l\ne m}F_{l}(x_{m})=\frac{1}{\sigma}y_{m}\exp\left(-y_{m}\left\{ 1+\sum_{l\ne m}e^{\frac{\mu_{l}-\mu_{m}}{\sigma}}\right\} \right)
\]
Now, by changing variable under the integration from $x_{m}$ to $y_{m}$,
we have 
\begin{eqnarray*}
P(e=c_{m}) & = & \sigma\int_{0}^{\infty}P(y_{m})\prod_{l\ne m}F_{l}(y_{m})\frac{1}{y_{l}}dy_{m}\\
 & = & \int_{0}^{\infty}\exp\left(-y_{m}\left\{ 1+\sum_{l\ne m}e^{\frac{\mu_{l}-\mu_{m}}{\sigma}}\right\} \right)dy_{m}\\
 & = & \frac{1}{1+\sum_{l\ne m}e^{\frac{\mu_{l}-\mu_{m}}{\sigma}}}\\
 & = & \frac{e^{\mu_{m}/\sigma}}{\sum_{l}e^{\mu_{l}/\sigma}}
\end{eqnarray*}

\subsubsection{Gumbel Distribution for Rank \label{sub:Gumbel-Distribution-Rank}}

We now extend the case of categorical evidences rank evidences. Again
we maintain one Gaussian variable per category. Without loss of generality,
for a particular rank $\pi$ we assume that we must ensure that $x_{1}>x_{2}>...>x_{D}$.
This is equivalent to 
\[
\left[x_{1}>\max_{l>1}x_{l}\right]\cap\left[x_{2}>\max_{l>2}x_{l}\right]\cap...\cap\left[x_{D-1}>x_{D}\right]
\]
The probability of this is essentially 
\[
P\left(\left\{ e_{m}=m\right\} _{m=1}^{D}\right)=P(e_{1}=1)\prod_{m>2}P\left(e_{l}=m\mid\left\{ e_{d}=l\right\} _{l=1}^{m-1}\right)
\]

In words, this offers a stagewise process to rank categories: first
we pick the best category, the pick the second best from the remaining
categories and so on (see also \cite{fligner1988multistage}).

The probability of picking the best category out of a subset is already
given in Appendix~\ref{sub:Gumbel-Distribution-Cat}: 
\begin{eqnarray*}
P(e_{1}=1) & = & \frac{e^{\mu_{1}/\sigma}}{\sum_{l\ge1}e^{\mu_{l}/\sigma}}\\
P\left(e_{l}=m\mid\left\{ e_{d}=l\right\} _{l=1}^{m-1}\right) & = & \frac{e^{\mu_{m}/\sigma}}{\sum_{l\ge m}e^{\mu_{l}/\sigma}}
\end{eqnarray*}
This gives us the Plackett-Luce model \cite{luce1959individual,plackett1975analysis}
as mentioned in \cite{stern1990models}.

\subsection{Global Attitude: Sample Questions}
\begin{itemize}
\item \textbf{Q4} (\emph{Ordinal}): {[}...{]} how would you describe the
current economic situation in (survey country) \textendash{} \{\emph{very
good, somewhat good, somewhat bad, or very bad}\}? 
\item \textbf{Q11a} (\emph{Binary}): How do you think people in other countries
of the world feel about China? -- \{\emph{like, disliked}\}? 
\item \textbf{Q35,35a} (\emph{Category-ranking}): Which one of the following,
if any, is hurting the world's environment the most/second-most \{\emph{India,
Germany, China, Brazil, Japan, United States, Russia, Other}\}? 
\item \textbf{Q76} (\emph{Continuous}): How old were you at your last birthday? 
\item \textbf{Q85} (\emph{Categorical}): What is your current employment
situation \{\emph{A list of employment categories}\}? 
\end{itemize}

\subsection{Other Supporting Materials}

\subsubsection{Laplace Approximation \label{sub:Laplace-Approximation}}

Laplace approximation is the technique using a Gaussian distribution
to approximate another distribution. For the univariate case, assume
that the original density distribution has the form 
\[
P(x)\propto\exp\left\{ -f(x)\right\} 
\]
First we find the mode $\mu$ of $P(x)$ or equivalently the minimiser
of $f(x)$ given it exists. Then we apply Taylor's expansion 
\[
f(x)\approx f(\mu)+f''(\mu)\frac{\left(x-\mu\right)^{2}}{2}
\]
The Gaussian approximation has the form 
\[
P^{*}(x)\propto\exp\left\{ -f''(\mu)\frac{\left(x-\mu\right)^{2}}{2}\right\} 
\]
where $1/f''(\mu)$ is the new variance.

\subsubsection{Some Properties of the Truncated Normal Distribution \label{sub:Some-Properties-of-Truncated-Normal}}

For a normal distribution $P(x|\mu,\sigma)$ of mean $\mu$ and standard
deviation $\sigma$ truncated from both sides, i.e., $\alpha<x<\beta$,
the new density reads

\[
\bar{P}_{[\alpha,\beta]}(x\mid,\mu,\sigma)=\frac{Q(x^{*})}{\sigma\left[\Phi(\beta^{*})-\Phi(\alpha^{*})\right]}
\]
where 
\begin{eqnarray*}
x^{*} & = & \frac{x-\mu}{\sigma};\quad\alpha^{*}=\frac{\alpha-\mu}{\sigma};\quad\beta^{*}=\frac{\beta-\mu}{\sigma}
\end{eqnarray*}
and $Q(\cdot)$ and $\Phi(\cdot)$ are the probability density function
and the cumulative distribution of the standard normal distribution,
respectively. In particular, we are interested in the mean of the
distribution $\bar{P}_{[\alpha,\beta]}$: 
\[
\bar{\mu}=\mu_{i}+\sigma\frac{Q(\alpha^{*})-Q(\beta^{*})}{\Phi(\beta^{*})-\Phi(\alpha^{*})}
\]
Some special cases: 
\begin{itemize}
\item When $\alpha=\beta$, this distribution reduces to the Dirac's delta. 
\item When $\alpha=-\infty$, we have a one-sided truncation from above
since $\Phi(\alpha^{*})=0$. 
\item When $\beta=+\infty$, we obtain a one-sided truncation form below
since $\Phi(\beta^{*})=0$. 
\end{itemize}

\end{document}